%% file: root.tex
\documentclass[letterpaper, 10 pt, conference]{ieeeconf}  

\IEEEoverridecommandlockouts                              

\usepackage{math}
\usepackage{fontawesome5}
\usepackage{times}
\usepackage{multicol}
\usepackage[colorlinks,bookmarks=true]{hyperref} 

\usepackage{cite}  

\usepackage{xurl}
\usepackage{caption}
\usepackage[table]{xcolor}

\usepackage{etoolbox}
\makeatletter
\patchcmd{\thebibliography}{\section*{\refname}}{}{}{} 
\makeatother

\definecolor{ours}{RGB}{44, 150, 44} 
\definecolor{vint}{RGB}{31, 119, 180} 
\definecolor{nomad}{RGB}{255, 127, 14} 
\definecolor{gnm}{RGB}{148, 103, 189} 
\definecolor{hst}{RGB}{23, 190, 207} 
\definecolor{gt}{RGB}{214, 39, 40} 
\definecolor{goal}{RGB}{255, 215, 0} 

\newcommand{\ours}{\textsc{HumAIN}}

\title{\LARGE \bf
\textsc{HumAIN}: Human-Aware Implicit Social Robot Navigation
}

\author{
Daeun Song$^{1}$, Nhat Le$^{2}$, Jeffrey Chen$^{3}$, Mohammad Nazeri$^{2}$, Amirreza Payandeh$^{2}$, \\Rohan Chandra$^{3}$, Reuth Mirsky$^{4}$, Ross Mead$^{5}$, Ling Xiao$^{6}$, Xuesu Xiao$^{2}$%
\thanks{$^{1}$Ewha Womans University, Korea. {\tt songd@ewha.ac.kr}; $^{2}$George Mason University, USA. {\tt \{nle47, mnazerir, apayande, xiao\}@gmu.ac.kr}; $^{3}$University of Virginia, USA. {\tt \{fyy2wsm, rohanchandra\} @virginia.edu}; $^{4}$Tufts University, USA. {\tt reuth.mirsky@tufts.edu}; $^{5}$Semio, USA. {\tt ross@semio.ai}; $^{6}$Hokkaido University, Japan. {\tt ling@ist.hokudai.ac.jp}}%
}

\begin{document}

\bstctlcite{IEEEexample:BSTcontrol}

\maketitle
\thispagestyle{empty}
\pagestyle{empty}

\input{content/0_abstract}

\input{content/1_intro}
\input{content/2_related}

\input{content/3_method}

\input{content/4_experiment}
\input{content/5_conclusion}

\input{content/6_ack}




\bibliographystyle{IEEEtran}
\bibliography{IEEEabrv, ref}

\end{document}

%% file: content/0_abstract.tex



\begin{abstract}

Effective social robot navigation requires sensitivity to human behavior, often revealed through subtle skeletal cues like gait and orientation. 
We present Human-Aware Implicit Social Robot Navigation (\ours), a novel framework that fuses implicit social cues directly into the planning loop via knowledge distillation. We first employ a transformer-based teacher model that fuses rich multi-modal inputs, including historic images, skeletal keypoints, robot state, and a robot's target goal, to learn robust, human-aware representations for the robot's future trajectory planning. To enable real-time deployment, we then distill this knowledge into a lightweight student model. By optimizing for both trajectory reconstruction and latent feature alignment with the teacher, the student learns to infer complex social dynamics from minimal inputs. 
Bridging the prediction–planning gap with an efficient distilled architecture, our method enables robots to reason about human behavior in a manner that is adaptive, robust, and socially compliant. 
We validate \ours{} through extensive experiments, where it improves trajectory prediction metrics by an average of 29.8\% across all metrics compared to state-of-the-art baselines. These results highlight the benefit of using implicit, whole-body cues to achieve human-like navigation awareness on resource-constrained platforms.


\end{abstract}

%% file: content/1_intro.tex
\section{Introduction} 

Social robot navigation is a critical area of robotics, driven by the growing demand for robots to operate autonomously in human-populated environments, from hospitals~\cite {diligent2020_spectrum} to public spaces~\cite{starship2021}. Unlike navigation in static or industrial domains, social navigation requires sensitivity to human behaviors and comfort~\cite{mirsky2024conflict}. To navigate fluently, a robot must go beyond collision avoidance and reason about others' intended paths, such as differentiating a pedestrian continuing straight from one preparing to turn \cite{mirsky2024recognition}. Human motor control studies suggest that pedestrians implicitly reveal navigational goals through body pose, head orientation, and gait~\cite{imai2001interaction}. For example, body pose can often indicate whether a person is walking or standing, and their intended walking direction. 

However, despite the richness of these signals, many motion planning approaches oversimplify humans as low-dimensional 2D points or moving circles~\cite{ferrer2013robot}. They typically react only to position and velocity using distance keeping~\cite{ferrer2013social} or proxemic rules~\cite{mead2017autonomous, charalampous2017recent}. While efficient, such representations discard body language, making it difficult to perceive preparatory cues. As a result, pedestrian turns can appear more difficult to anticipate.

\begin{figure}[t]
\centering
\includegraphics[width=\linewidth]{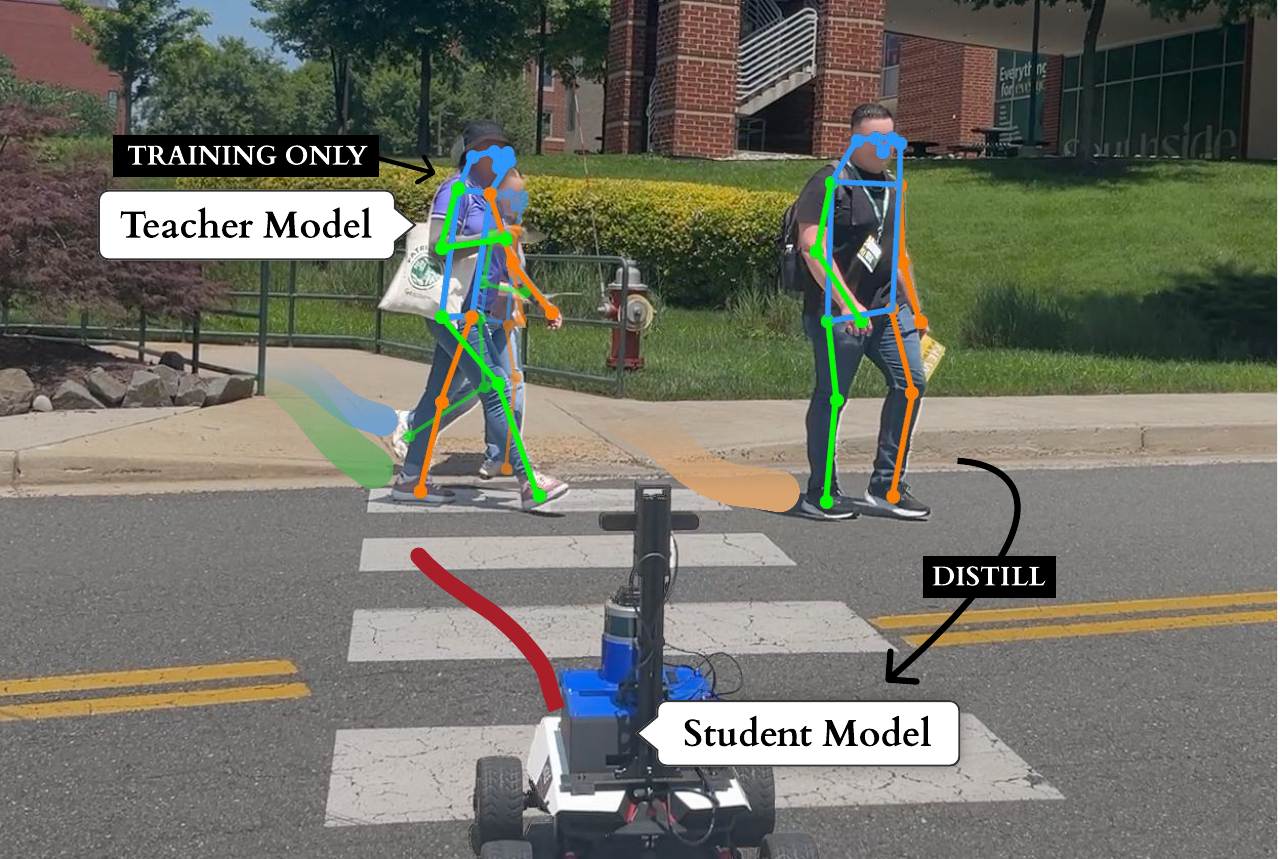} \vspace{-1.0em} 
\caption{\ours\ distills a privileged knowledge from a teacher model into a student model. While the teacher uses these keypoints to plan robot trajectories, the student learns to infer social cues directly from RGB input to plan safe, socially compliant trajectories end-to-end.}
\vspace{-1.5em}
\label{fig:cover}
\end{figure} 

While human trajectory prediction methods have leveraged pose to improve forecasting~\cite{alahi2016social, salzmann2023hst}, integrating these predictions into robot trajectory planning remains challenging. Many systems operate in a decoupled manner by forecasting a human path and then planning a robot trajectory to avoid it~\cite{poddar2023crowd}. This modular approach treats forecasts as fixed constraints and often yields overly conservative maneuvers or inefficient detours.

In contrast, humans implicitly use social cues such as body pose, hip orientation, gaze, and gait to infer others' short-term motion and adjust their own trajectories accordingly~\cite{kitagawa2021human}. 
Yet robots often lack the sensors or compute budget to track such cues reliably at deployment time, which creates a gap between the rich information needed for social reasoning and the limited sensorimotor inputs available at deployment. A natural solution is to learn representations with a computationally powerful teacher that uses privileged information during training as scaffolding, then distill this knowledge into an efficient onboard student that reasons from raw observations.

Inspired by this intuition, we propose a Teacher--Student framework that incorporates implicit social cues into robot planning (Fig.~\ref{fig:cover}). We train an Oracle Teacher with privileged multi-modal data, including skeletal keypoints, to learn a rich latent representation of human behavior. We then distill this representation into a lightweight sensorimotor student that operates only on raw images and a goal, enabling socially compliant behavior without explicit pose tracking.

\noindent {\bf Main Results:} 
We present \textbf{\ours}, \textbf{Hum}an-\textbf{A}ware \textbf{I}mplicit social robot \textbf{N}avigation, a social navigation approach that incorporates whole-body pose cues via response-based knowledge distillation~\cite{gou2021knowledge}. We employ a two-stage training strategy. A Teacher learns multi-modal representations from images and skeletal keypoints. A Student is trained to match the Teacher latent representation and imitate its behavior using only raw images. This process enables leveraging skeletal supervision during training while remaining efficient and sensor-agnostic at deployment.

The main contributions of this work are as follows:
\begin{itemize} 
    \item \ours, an end-to-end framework for social navigation that leverages whole-body pose via skeletal keypoints as an implicit cue for robot trajectory planning. 
    \item A distillation strategy that aligns a Student with a privileged Teacher, enabling inference of complex social cues from raw visual inputs. 
    \item Extensive quantitative and qualitative evaluations and ablations showing the impact of pose cues on trajectory planning and overall navigation performance. \ours{} improves social navigation performance by an average of 29.8\% over state-of-the-art baselines and all metrics.
\end{itemize}

%% file: content/2_related.tex
\begin{figure*}[ht]
\centering
\includegraphics[width=0.9\linewidth]{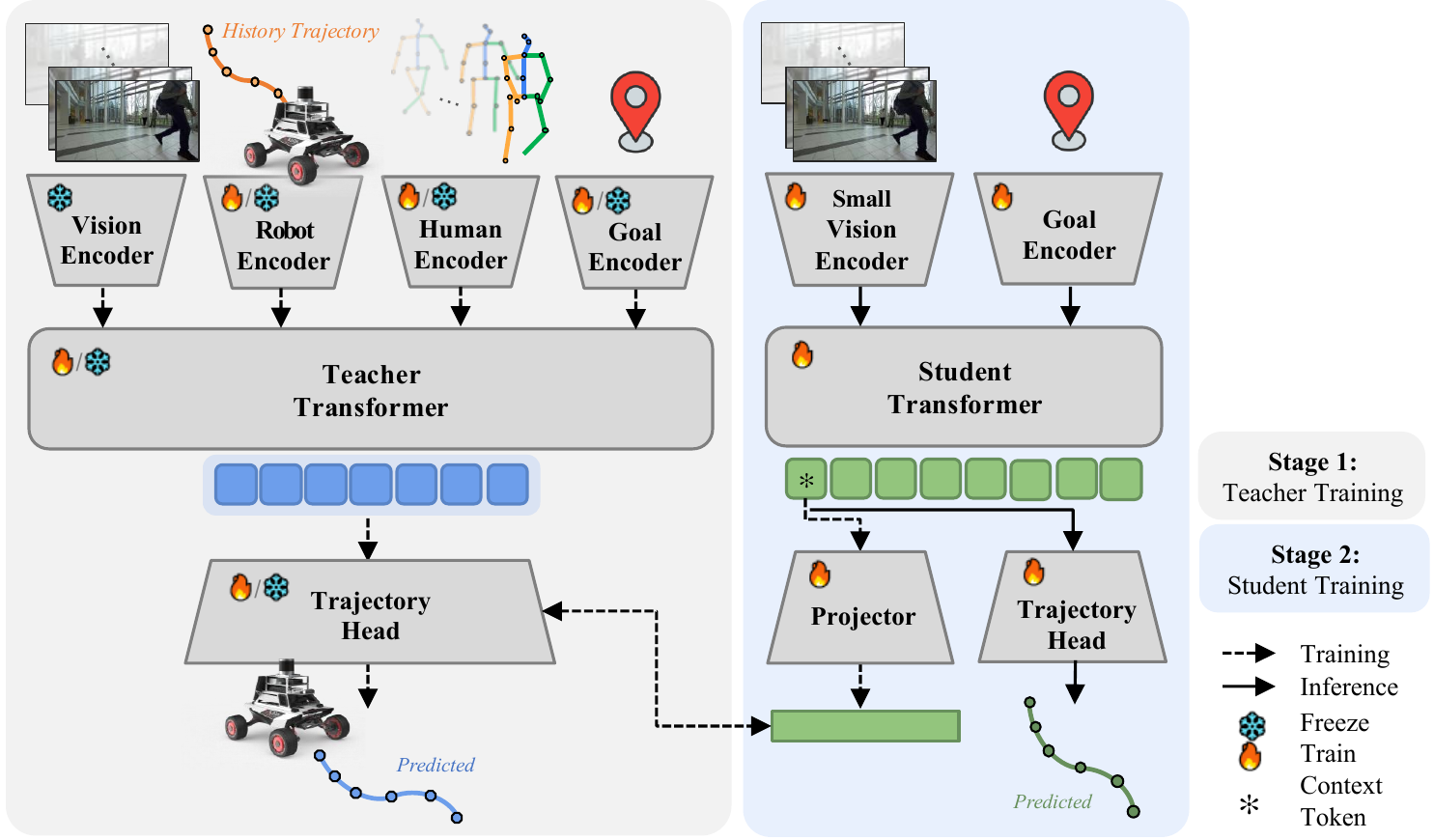}
\caption{\textbf{\ours{} Architecture.} Two-stage framework. The Oracle Teacher uses privileged multi-modal inputs including dense keypoints and trajectory history to learn social representations and predict trajectories. The Student is distilled to navigate from raw images and goal by matching the frozen Teacher latent features. Solid lines denote inference. Dotted lines denote training-only distillation.}
\vspace{-1.5em}
\label{fig:architecture}
\end{figure*}

\section{Related Work}
\label{sec:related_work}

We review work on social robot navigation, human trajectory prediction, and representation learning for navigation.

\subsection{Social Robot Navigation}

Classical social navigation methods based on proxemics~\cite{mead2017autonomous, charalampous2017recent}, social force models~\cite{ferrer2013social}, cost maps~\cite{mead2017autonomous}, and handcrafted rules~\cite{ mirsky2024recognition} offer safety and explainability~\cite{raj2024}, but often require substantial tuning and heuristics to handle complex scenarios~\cite{xiao2022appl, wang2021appli}. Learning-based approaches reduce manual design via imitation learning~\cite{nguyen2023toward, karnan2022voila} and reinforcement learning~\cite{wang2023navistar, faust2018, chandra2025multi}, but can be less interpretable and brittle in edge cases~\cite{raj2024}. Prior work incorporated orientation or gait cues with handcrafted perception~\cite{ratsamee2013social, narayanan2020proxemo}. More recently, VLM-based methods~\cite{vlmsocialnav, narasimhan2025olivia, payandeh2024social} provide semantic grounding, but they are computationally heavy and sensitive to prompts and domain shifts~\cite{wang2025maction}, and they often require costly language supervision. These limitations motivate our approach, which learns implicit human cues via representation learning and distillation without handcrafted rules or expensive labels.

\subsection{Human Trajectory Prediction}

Human trajectory prediction is closely related to social navigation, since anticipating motion supports safe and natural robot behaviors. Early generative approaches modeled interactions with recurrent networks or GANs, such as Social LSTM~\cite{alahi2016social} and Social GAN~\cite{gupta2018social}, which capture multi-agent dependencies. More recently, Transformer-based and graph-based methods have advanced the field by explicitly reasoning over spatial--temporal relations. For instance, Human Scene Transformer~\cite{salzmann2023hst} models multi-human interactions with attention, SocialPose~\cite{gao2025social} leverages pose features for forecasting in crowds, Social-Transmotion~\cite{saadatnejad2023social} exploits human poses and bounding boxes for trajectory forecasting, and UPTor~\cite{nilavadi2025uptor} jointly predicts 3D body pose dynamics and trajectories for human--robot interaction scenarios. Some works integrate predictors into MPC for robot deployment~\cite{poddar2023crowd}, yet strong prediction accuracy does not necessarily yield socially compliant navigation behaviors.

\subsection{Representation Learning for Navigation}

Representation learning~\cite{bengio2013representation}, often achieved through self-supervised learning (SSL), enables robots to acquire robust features without costly labels. In navigation, representation learning has been applied in diverse ways. Prior work includes vision--action pretraining~\cite{nazeri2024vanp}, 
language-weak supervision~\cite{narrate2nav}, and trajectory-centric SSL~\cite{nazeri2025verticoder}. Large-scale supervised policies such as ViNT~\cite{vint} and NoMaD~\cite{nomad} further improve transfer, but most still focus on low-level control, and do not explicitly capture social cues.

Overall, recent work has shown that incorporating human body pose improves trajectory prediction, demonstrating the value of pose and gait cues for modeling human motion~\cite{gao2025social, nilavadi2025uptor}. However, these advances have not yet translated into social robot navigation, where many methods focus on geometric safety, semantic grounding, or general representations from images. Our approach extends the use of skeletal keypoints from trajectory prediction to navigation, and employs SSL and distillation to embed implicit human cues directly into robot planning.


%% file: content/3_method.tex

\section{Approach}
\label{sec:approach}

In this section, we present details of our proposed \ours{}, a hierarchical learning framework for socially compliant navigation. We first define the problem as a trajectory planning problem, given the observation and goal. We then provide an overview of our architecture, followed by detailed descriptions of the multi-modal representation learning in the Teacher model and the knowledge distillation strategy used for the student model. 

\subsection{Problem Definition}

We formally define social robot navigation as a goal-conditioned trajectory planning problem. The robot receives an observation history $O \in \co^{T_\textrm{obs}}$ and a local navigation goal $g \in \cg$, where $\co$ is the observation space, $T_\textrm{obs}$ the observation horizon, and $\cg$ the goal space. The objective is to generate a future trajectory $\hat{Y}\in \cy^{T_\textrm{pred}}$, where $\cy$ represents the robot's trajectory space and $T_\textrm{pred}$ is the prediction horizon. The trajectory must progress toward $g$ while adhering to social norms. To address the challenge of learning social compliance without manual cost engineering, we formulate this process under two distinct information regimes:

\vspace{5pt}

\noindent \textbf{Privileged Representation Learning (Teacher):} 
The primary objective of the first stage is to learn a robust latent representation that encapsulates complex social dynamics from privileged observation data. During training, the system has access to an oracle observation set $O^\textrm{teacher}_t=(I_t,R_t,H_t)$, comprising egocentric RGB images $I_t \in \ci$, the robot's state history $R_t \in \crr$, and privileged human states $H_{t,m} \in \ch$ for up to $M$ humans. 
Specifically, we define the state of each human $m$ at time $t$ as a tuple $H_{t,m} = (\mathbf{k}_{t,m}, \mathbf{p}_{t,m}, \mathbf{d}_{t,m})$, comprising the 3D skeletal keypoints, the 2D position relative to the robot, and the 2D facing direction relative to the robot. 
We seek to learn a mapping function $F_T:(\ci \times \crr \times \ch)^{T_\textrm{obs}} \times \cg \rightarrow \cy^{T_\textrm{pred}}$ parameterized by $\theta_\textrm{teacher}$. The Teacher leverages this full spectrum of spatio-temporal data as scaffolding to generate a future trajectory $\hat{Y}_\textrm{teacher}$ that minimizes the prediction error against the ground truth $Y_\textrm{GT}$. 
Crucially, while the explicit output is the predicted trajectory, the learned latent representation $\cz_\textrm{teacher}$ captures social cues encoded within the skeletal data in $\ch$ along with other observations. This latent representation is later used as the distillation target for the Student.

\vspace{5pt}

\noindent \textbf{Sensorimotor Policy Distillation (Student):} The Student operates under realistic deployment constraints where human tracking data is unavailable or computationally prohibitive. The observation is restricted to the sensorimotor subset $O^\textrm{student}_t =  I_t$, containing only the visual history. We seek a deployment policy $F_S:\ci^{T_\textrm{obs}} \times \cg \rightarrow \cy^{T_\textrm{pred}}$, parameterized by $\theta_\textrm{student}$. Since the sensorimotor input space is a subset of the privileged input space, minimizing trajectory error alone is often insufficient to recover complex social behaviors. Therefore, the Student's objective is two-fold: (1) to generate a socially-compliant trajectory $\hat{Y}_\textrm{student}$ that matches the ground truth $Y_\textrm{GT}$, and (2) to align its internal belief state with the Teacher's privileged representation $\cz_\textrm{teacher}$. This dual objective forces the Student to implicitly reason about the missing social cues, such as human whole-body pose, solely from the available visual inputs. 

\subsection{Model Architecture}
In Fig.~\ref{fig:architecture}, we illustrate the overall architecture of \ours. Our framework consists of two Transformer-based networks: the Oracle Teacher and the Sensorimotor Student. 

The Oracle Teacher is a high-capacity model designed to process multi-modal inputs, including dense human keypoints and robot state history, to generate socially aware trajectories. Each modality is processed by a dedicated encoder to project heterogeneous inputs into a shared embedding space. These embeddings are then fused through a hierarchical, robot-centric attention pipeline. The resulting tokens are passed to the Trajectory Head, implemented as a Multilayer Perceptron (MLP), for trajectory prediction. 

The Sensorimotor Student is a lightweight model for real-time inference on mobile robots. It takes only a sequence of raw images and the navigation goal as input. These inputs are encoded by a lightweight vision encoder and a goal encoder, respectively, before being fused by the Student Transformer. A key architectural distinction lies in the token processing strategy. Unlike the Teacher, which utilizes the full sequence of Transformer output tokens, the Student extracts only a single global context token. This token is passed through a projector and a lightweight MLP head to regress to the final trajectory. This design forces the Student to compress complex social reasoning into a compact latent representation, which is explicitly aligned with the Teacher's privileged representation via knowledge distillation.

\subsection{Multi-Modal Representation Learning}
The primary goal of the first training stage is to learn the Teacher mapping function $F_T$ and, more importantly, to construct a regularized latent manifold $\mathcal{Z}_\textrm{teacher}$ that encapsulates high-fidelity social semantics. The Oracle Teacher processes the privileged observation sequence $O^\textrm{teacher}=\{(I_t,R_t,H_t)\}_{t=1}^{T_\textrm{obs}}$ through modality-specific encoders, followed by a hierarchical fusion pipeline composed of two Transformer-based modules that operate in sequence.

\vspace{5pt}\noindent \textbf{Modality-Specific Encoding:}
Each modality is projected into a shared embedding dimension $D$ to form tokens. 
\begin{itemize}
    \item \textit{Visual Tokens:} A frozen pretrained backbone (e.g., DINOv3~\cite{dinov3}) encodes egocentric RGB frames. We use all tokens, not only the global token, to preserve spatial context. 
    The resulting visual tokens are defined as $\mathbf{Z}^I \in \mathbb{R}^{T \times N \times D}$, where $N$ includes patch, class, and register tokens per frame.
    \item \textit{Robot Trajectory Tokens:} The robot kinematic history (e.g., waypoints) is encoded by an MLP into  $\mathbf{Z}^R \in \mathbb{R}^{T \times 1 \times D}$. 
    \item \textit{Goal Token:} The goal is projected into a single token $\mathbf{z}^g \in \mathbb{R}^D$, providing the directional context.
\end{itemize}

\vspace{5pt}\noindent \textbf{Body-Part-Aware Human Encoder:}
To extract expressive social cues, the human encoder partitions the $J$ skeletal keypoints of $M$ humans into a set of semantic body parts, $b \in \mathcal{B} = \{\text{Head, Torso, Arms, Legs}\}$. For each human $m$ at time $t$, the skeletal keypoints $\mathbf{k}_{t,m}$ of each group are processed by a part-specific MLP to extract part-level features. We add a learnable part-type embedding to each of the four outputs, explicitly tagging each vector as belonging to a specific body region. These tagged tokens are then concatenated with the human’s relative position $\mathbf{p}_{t,m}$ and facing direction $\mathbf{d}_{t,m}$, grounding the postural representation in the robot's egocentric frame. Finally, a fusion MLP projects this combined vector into a single $D$-dimensional human token $\mathbf{z}^H_{t,m}$, which encapsulates the human's posture and intent while preserving a fixed structural hierarchy. The resulting output is a sequence $\mathbf{Z}^H \in \mathbb{R}^{T \times M \times D}$.



\vspace{5pt}\noindent \textbf{Hierarchical Robot-Centric Fusion:} 
We employ a hierarchical fusion strategy that mirrors the cognitive stages of navigation: social awareness, scene understanding, and goal-directed reasoning. The process is structured into three distinct stages: 
\begin{itemize}
    \item \textit{Social-Robot Interaction:} Inspired by the Human Scene Transformer (HST)~\cite{salzmann2023hst}, we model the joint dynamics of the robot and humans. While we adopt the dual-stage temporal and social attention paradigm from HST, we extend this by enabling bidirectional robot-human attention. This ensures the robot’s latent state is contextualized by human intent while explicitly modeling the robot's own influence on surrounding agents.
    \item \textit{Vision-Robot Grounding:} The social-aware robot tokens from the preceding interaction layer act as queries in a cross-attention mechanism over the visual token sequence $\mathbf{Z}^I$. This attention is strictly time-aligned: the robot state at time $t$ queries only the visual features from the corresponding frame. This ensures temporal consistency, allowing the model to ground its social reasoning within the physical constraints of the environment, such as obstacles or navigable pathways.
    \item \textit{Goal-Robot Reasoning:} In the final stage, the enriched robot tokens are concatenated with the time-invariant goal token $\mathbf{z}^g$. A self-attention block facilitates a global reasoning step where temporal social-visual features are synthesized with the destination target to produce the final decision-making tokens.
\end{itemize}
By decoupling these interactions into a hierarchy, the model avoids the dilution of signal common in massive flat sequences and ensures that the robot remains the central agent throughout the reasoning process.

\vspace{5pt}\noindent \textbf{Trajectory Planning and Objective Function:}
The Transformer Encoder outputs a sequence of contextualized tokens. After hierarchical fusion, we obtain a sequence of $T+1$ tokens, comprising $T$ temporally-contextualized robot tokens and one goal token. We flatten this sequence into a single vector of dimension $(T_{obs}+1)\times D$, which serves as the input to the Trajectory Head. 
We pass this vector to a Trajectory Head, composed of a 4-layer MLP. This head maps high-dimensional scene features to the robot's future path. The first two layers perform initial feature expansion and processing, which are then followed by a critical third layer, which later serves as the distillation target. This layer produces the latent representation $\mathcal{Z}_\textrm{teacher} \in \mathbb{R}^{D}$ . This vector serves as a condensed semantic embedding, which encapsulates the high-fidelity social semantics extracted from skeletal and visual data. Finally, the fourth layer functions as the output head, generating the coordinate ($\hat{y} \in \mathbb{R}^2$) sequence for the predicted future trajectory $\hat{Y} \in \mathbb{R}^{T_\textrm{pred} \times 2}$. 
The training objective is to minimize the supervised trajectory prediction error. We employ a Mean Squared Error (MSE) loss between the predicted trajectory $\hat{Y}$ and the ground truth future trajectory $Y_\textrm{GT}$. 
By minimizing this loss, the Teacher model learns to optimize the latent manifold $\mathcal{Z}_\textrm{teacher}$ to be maximally predictive of socially compliant robot motion.

\subsection{Knowledge Distillation}

In this stage, the Student model operates under deployment constraints where human skeletal tracking is unavailable. The Student receives only the visual history $\I$ and the navigational goal $g$ to plan the future trajectory $\hat{Y}_\textrm{student}$. To ensure real-time performance onboard mobile robots, we replace the Teacher's heavy backbone with a lightweight, trainable convolutional neural network (e.g., ResNet~\cite{he2016deep}). The Student transformer model maintains a similar structural flow, mapping visual tokens and the goal to a fused context. However, the internal Transformer and MLP layers are scaled down to reduce computational overhead for onboard real-time inference. We construct the input sequence by processing the available modalities into a streamlined token set. The visual history is encoded into $T$ tokens, which are concatenated with a single token representing the goal embedding.

\vspace{5pt}\noindent \textbf{Feature Distillation Strategy:} 
After the Student Transformer fuses the visual and goal embeddings, we extract \textit{only} the output of a learnable CLS token, which is prepended to the input sequence and aggregates information from all other tokens through self-attention. This forces the model to compress the entire scene's social affordances into a compact belief state $z_\textrm{student}$. 
While inherently lossy, this compression is essential for low-latency inference. The burden of ensuring this single token captures complex social dynamics is handled by the distillation process. 
The distillation process leverages the frozen Teacher model to guide this compression. We extract the Teacher's latent representation $\cz_\textrm{teacher}$ from the third layer of its trajectory MLP, which encapsulates high-level reasoning about social interactions. 
Since the Student's compact belief state $z_\textrm{student}$ may differ in dimensionality from the Teacher's latent space, we introduce a learnable projector $\phi$. This projector maps the Student's token into the Teacher's feature space, allowing for direct alignment.

\vspace{5pt}\noindent \textbf{Objective Function:} 
The Student is trained using a dual-objective loss function to implicitly consider the missing social cues, solely from visual inputs.
The first component is the Feature Alignment Loss, $\mathcal{L}_\textrm{feat}$, which minimizes the distance between the Student's embedding $\mathcal{Z}_\textrm{student}$ and the Teacher's latent representation $\mathcal{Z}_\textrm{teacher}$. By targeting the manifold extracted from the third layer of the Teacher's trajectory head using an MSE criterion, the Student learns to approximate the Teacher's privileged social reasoning. Simultaneously, the Student must maintain accuracy in the primary navigation task through a Trajectory Imitation Loss, $\mathcal{L}_\textrm{traj}$. This component applies an MSE loss between the Student's predicted path $\hat{Y}_\textrm{student}$ and the ground truth trajectory $Y_\textrm{GT}$. 
The final Student policy is optimized by minimizing a weighted combination of these losses:
\begin{equation}
    \mathcal{L}_\textrm{total} = \mathcal{L}_\textrm{traj} + \lambda \mathcal{L}_\textrm{feat}    
\end{equation}
This training regime ensures that even without access to skeletal data at test time, the Student's navigation remains socially compliant by mimicking the Teacher’s privileged understanding of the scene.

\begin{figure}[tb]
\centering
\begin{tabular}{cc} 
    \includegraphics[width=0.48\linewidth]{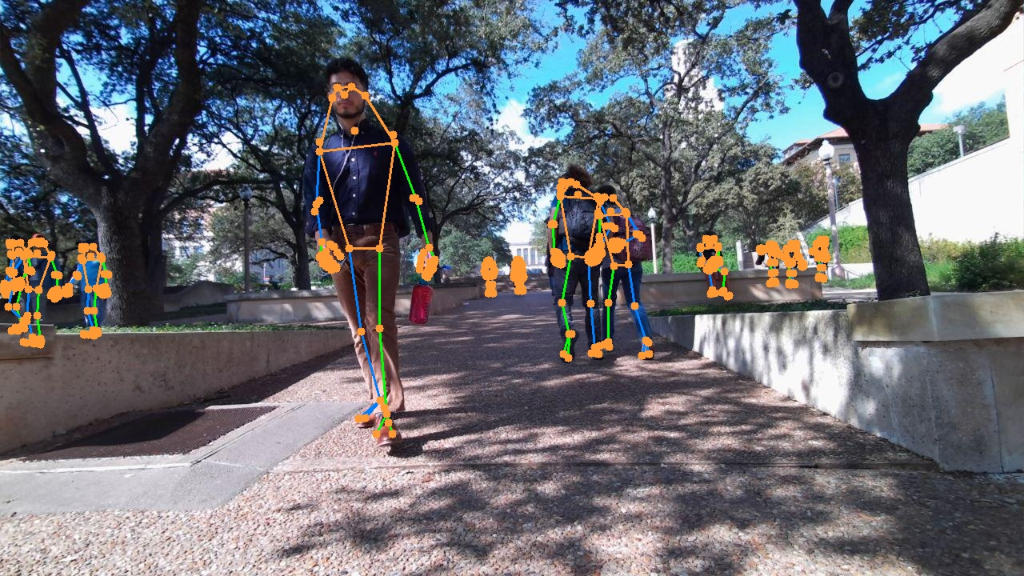} &
    \includegraphics[width=0.48\linewidth]{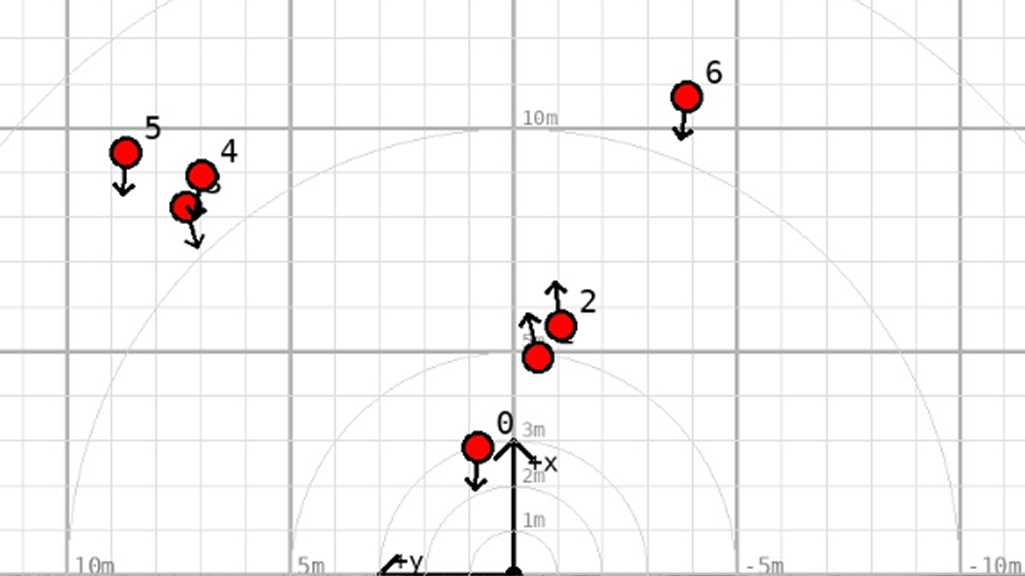}\\
    \includegraphics[width=0.48\linewidth]{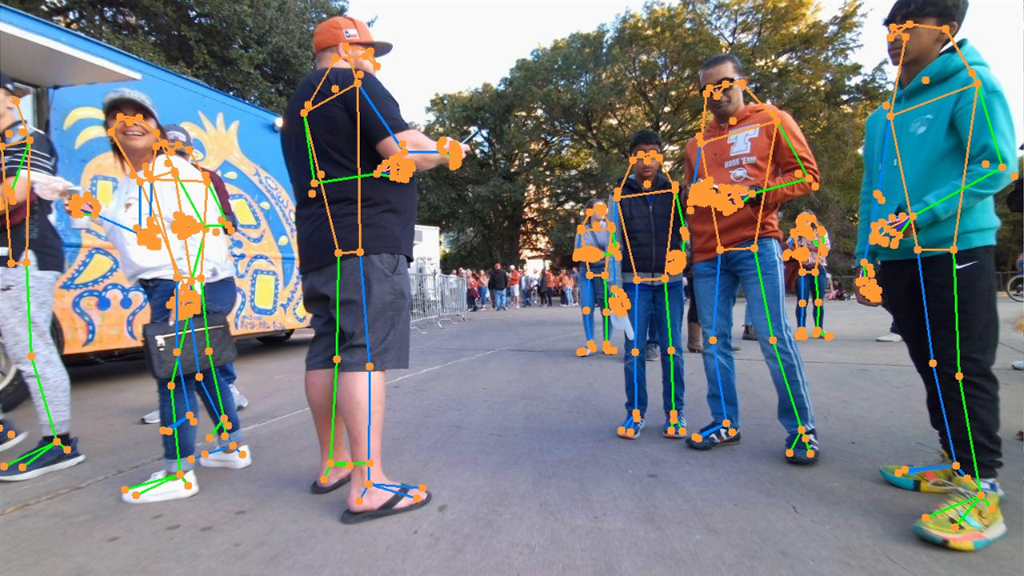}  &
    \includegraphics[width=0.48\linewidth]{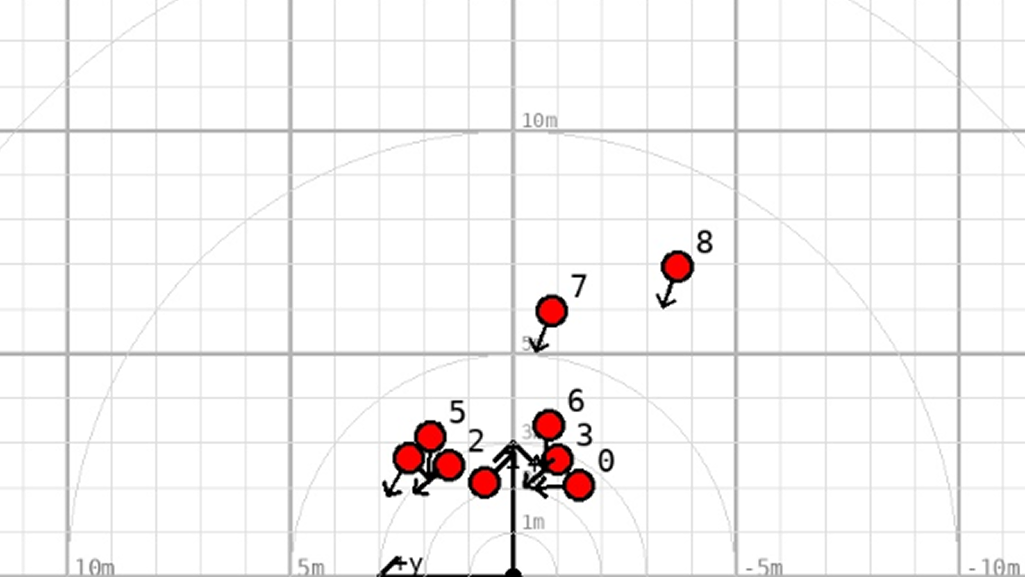} \\
    (a) Skeletal Keypoints & (b) Position and Direction \\
\end{tabular} 
\caption{\textbf{Visualization of the Human State Representation.} We employ SAM3DBody~\cite{sam3dbody} to extract social cues. (a) 3D skeletal keypoints overlaid on the robot's egocentric view. (b) A corresponding top-down perspective, where numbered nodes represent the relative 2D position, and arrows indicate the facing direction of each tracked human.
} \vspace{-1.0em}
\label{fig:data}
\end{figure}

%% file: content/4_experiment.tex
\setlength{\tabcolsep}{6pt}

\section{Experimental Results}

In this section, we outline the implementation details and datasets, followed by quantitative and qualitative results. 

\subsection{Experimental Setup}

We implement \ours{} with PyTorch and train our model on a single NVIDIA GeForce RTX 4070 Ti SUPER GPU with 16GB memory. Training is performed with the AdamW optimizer. We process the data at a frequency of 4Hz. The model utilizes an observation window of $T_{\text{obs}}=6$ frames (approximately 1.5s) to capture immediate historical context and forecasts a future trajectory over a prediction horizon of $T_{\text{pred}}=12$ frames (approximately 3.0s). 
For goal conditioning, the ground truth position at $15\mathrm{m}$ ahead is defined as the target goal.

\vspace{5pt}\noindent\textbf{Dataset:} We train our model on SCAND~\cite{karnan2022socially}, a social robot navigation dataset, where robots were teleoperated to demonstrate socially compliant behaviors in human-populated environments. To reduce ambiguity and noise, we filter out scenes without humans within $15\mathrm{m}$ or with minimal robot orientation change. The resulting dataset comprises about 12,000 samples for training, with an additional 2,000 samples reserved for testing. For fair comparison, we also evaluate on an independent Out-of-Distribution (OOD) dataset of about 2,000 samples collected by teleoperating an AgileX Scout Mini robot with similar socially compliant behaviors.

To obtain the privileged human-state inputs for the Teacher model, we apply SAM3DBody~\cite{sam3dbody} together with ByteTrack~\cite{zhang2022bytetrack} offline on RGB images to extract dense 3D human keypoints, as well as per-person 2D positions and facing directions used as training social cues (Fig.~\ref{fig:data}). We discard detailed hand and foot joints and retain 30 core keypoints corresponding to the head, torso, arms, and legs, supporting up to $M=10$ tracked humans per scene.


\vspace{5pt}\noindent\textbf{Baselines:} To validate \ours, we compare against four state-of-the-art methods. We include ViNT~\cite{vint}, NoMaD~\cite{nomad}, and GNM~\cite{gnm} as representative vision-only navigation baselines that share the same input modality as our Student. We also include HST~\cite{salzmann2023hst} as a modular baseline that first predicts human trajectories and then plans robot motion, enabling a direct comparison between prediction-then-planning pipelines and our end-to-end distillation approach. Unlike modular systems that require explicit human tracking at inference time, \ours\ runs from RGB images only.


To leverage HST~\cite{salzmann2023hst}'s probabilistic forecasts, we construct time-varying keep-out zones by inflating each predicted mean of human positions into an obstacle whose radius is proportional to the forecast variance. A Model Predictive Controller (MPC)~\cite{garcia1989model} planner then optimizes a robot trajectory that balances goal progress, smoothness, and control effort while avoiding these dynamic safety zones.



\vspace{5pt}

\noindent\textbf{Metrics:} Evaluating social navigation is challenging because factors like human comfort and perceived compliance are subjective~\cite{francis2025principles}. Since our data consist of expert teleoperation demonstrations, we evaluate performance by measuring divergence from the ground-truth trajectories, treating them as a proxy for socially compliant behavior. We report three standard metrics:
\begin{itemize}
    \item \textbf{Average Displacement Error (ADE)}: mean position error over all time steps.
    \item \textbf{Final Displacement Error (FDE)}: position error at the final prediction horizon.
    \item \textbf{Average Orientation Error (AOE)}: mean heading error relative to ground-truth orientation~\cite{liu2025citywalker}.
\end{itemize}



\setlength{\tabcolsep}{1pt}
\begin{figure*}[tb]
\centering
\begin{tabular}{ccccc}
\spheading{SCAND} &\includegraphics[width=0.24\linewidth]{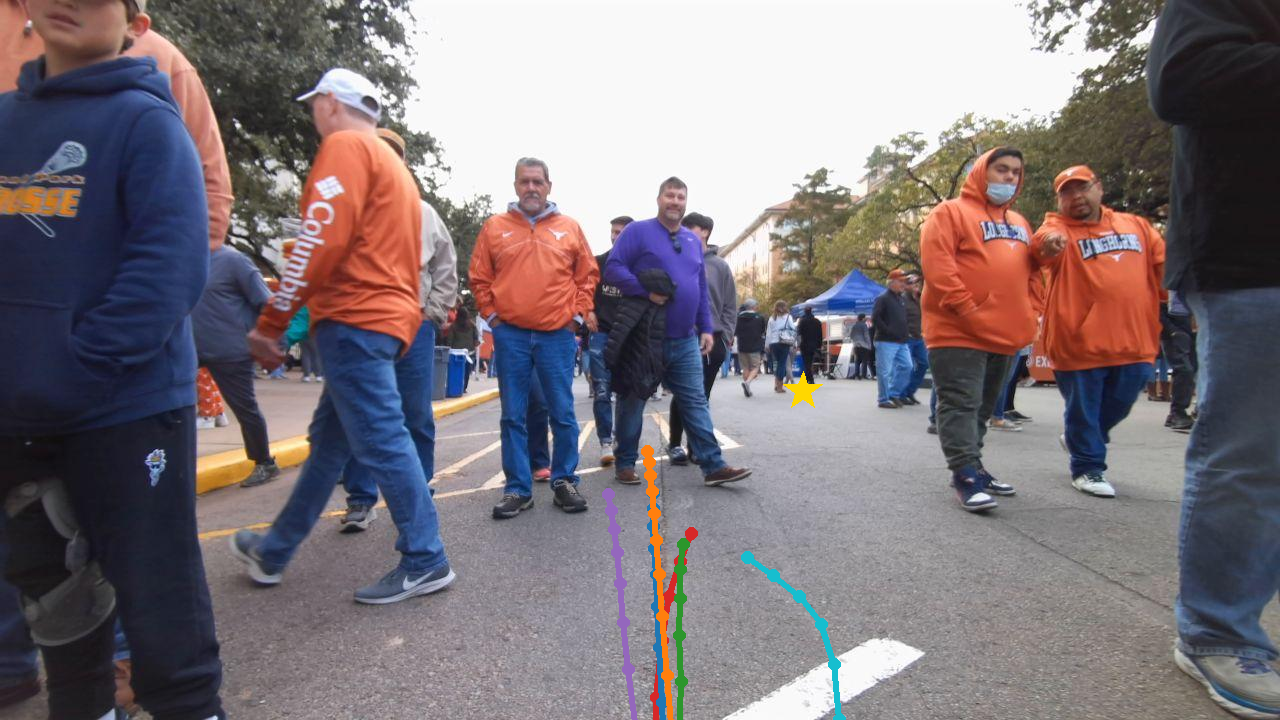} &
\includegraphics[width=0.24\linewidth]{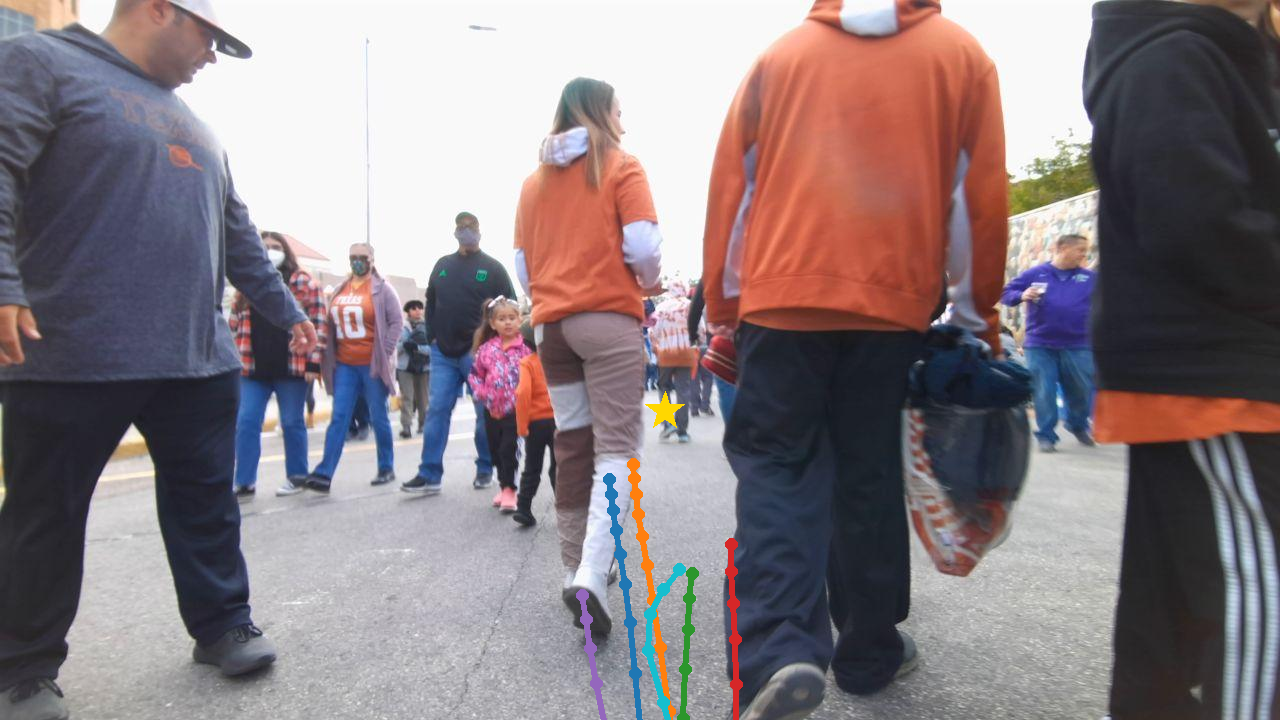} &
\includegraphics[width=0.24\linewidth]{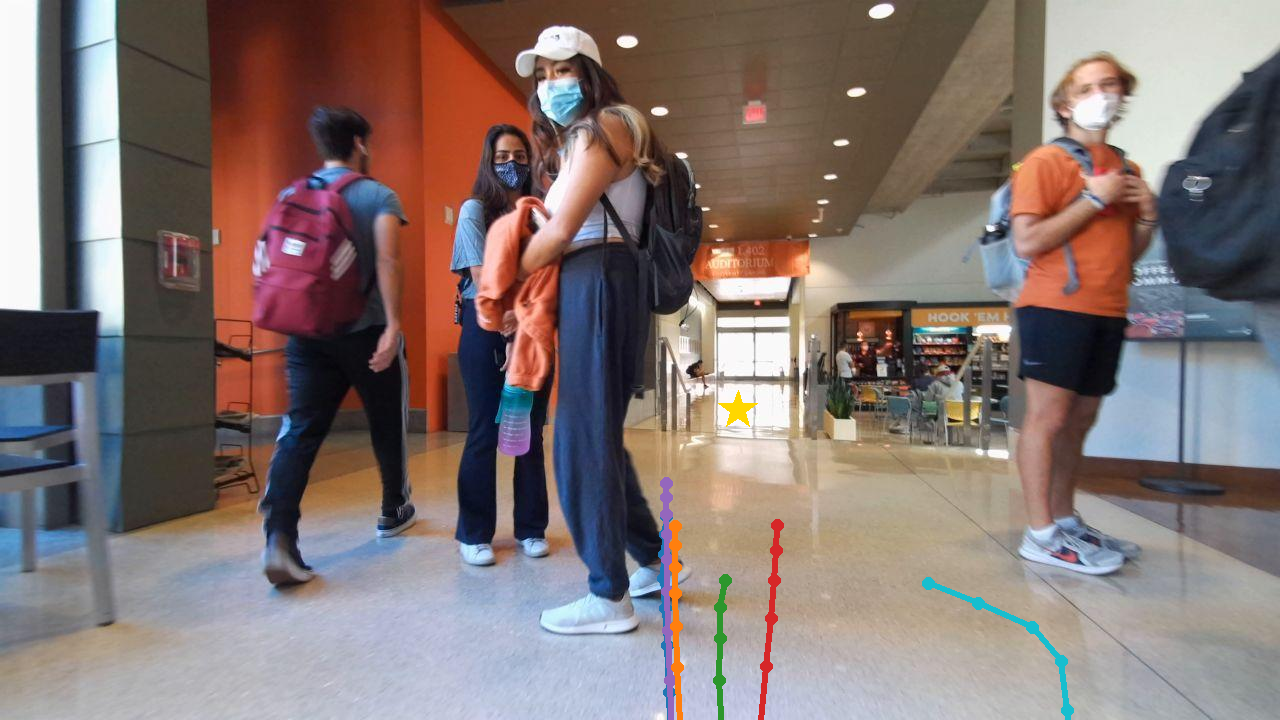} &
\includegraphics[width=0.24\linewidth]{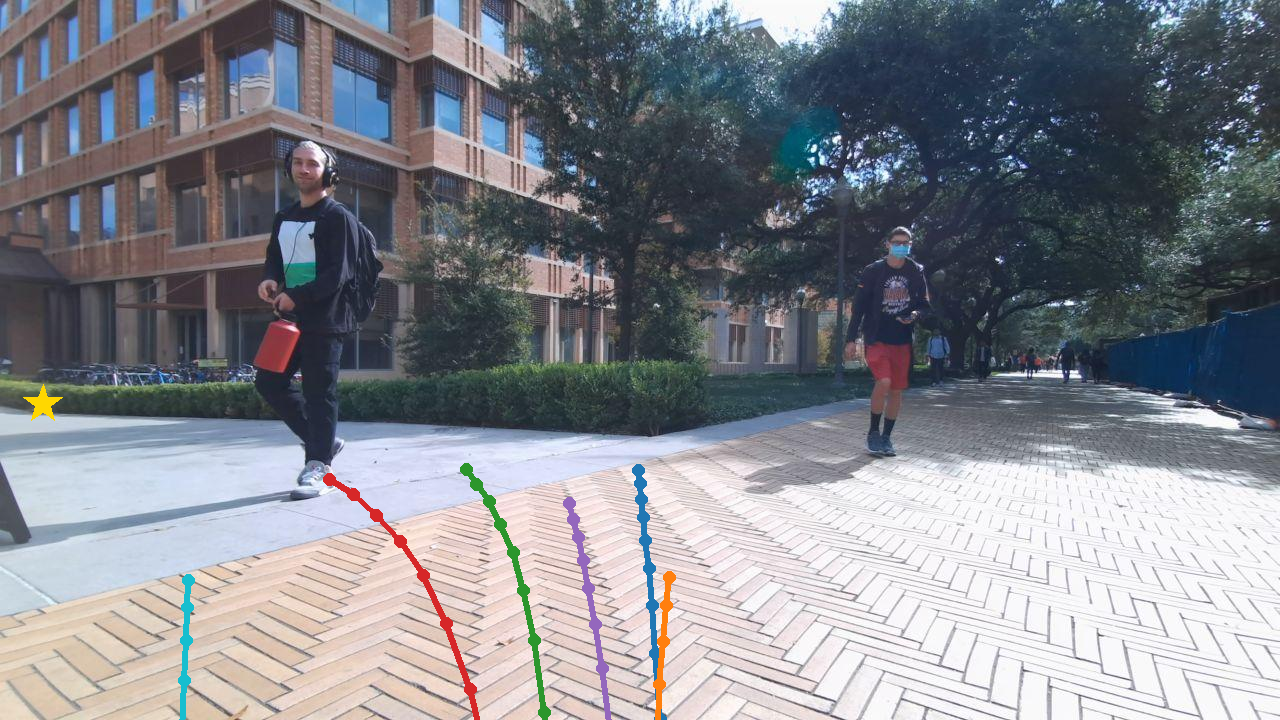} \\

\spheading{OOD} &\includegraphics[width=0.24\linewidth]{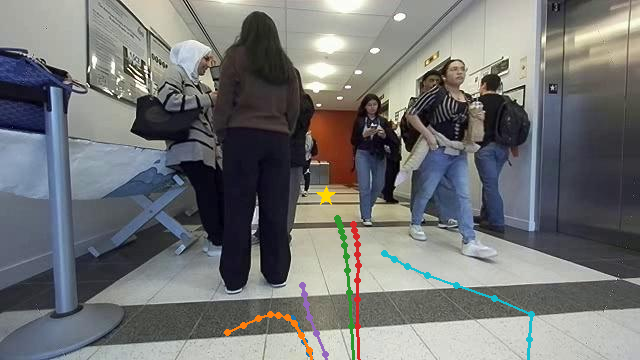} &
\includegraphics[width=0.24\linewidth]{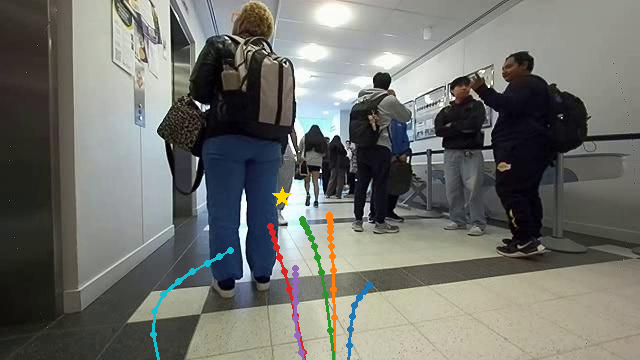} & 
\includegraphics[width=0.24\linewidth]{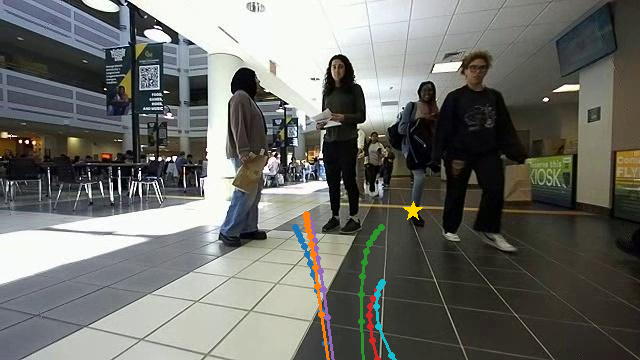} &
\includegraphics[width=0.24\linewidth]{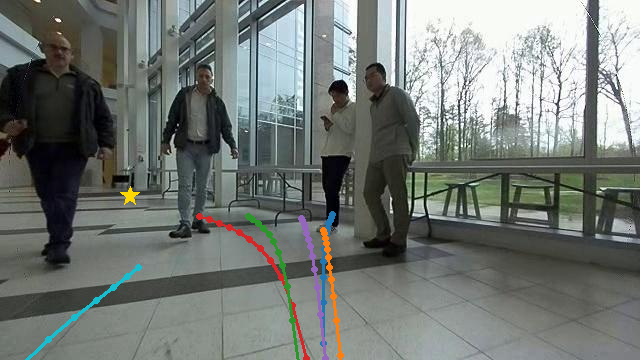} \\ 

\multicolumn{5}{c}{\small{
\textcolor{vint}{\thickline} ViNT 
\textcolor{nomad}{\thickline} NoMaD 
\textcolor{gnm}{\thickline} GNM 
\textcolor{hst}{\thickline} HST+MPC  
\textcolor{ours}{\thickline} \ours{} (Ours)}
\textcolor{gt}{\thickline} {Ground Truth}
\textcolor{goal}{$\bigstar$} {Goal}}

\end{tabular} 
\caption{\textbf{Qualitative Results:} Generated trajectories are projected onto the robot ego-centric view images using all the methods, ViNT~\cite{vint} in blue, NoMaD~\cite{nomad} in orange, GNM~\cite{gnm} in purple, HST~\cite{salzmann2023hst} with MPC~\cite{garcia1989model} in cyan, \ours{} in green, and the ground truth trajectory in red. The target goal is marked with a yellow star. The top row shows SCAND test dataset results, and the bottom row shows the results on the out-of-distribution dataset that we collected. }  \vspace{-1.5em}
\label{fig:qualitative}
\end{figure*}  
\setlength{\tabcolsep}{6pt} 

\setlength{\tabcolsep}{8pt}
\begin{table}[tb] 
\centering
\caption{\textbf{Quantitative Results}: We compare \ours{}\ with state-of-the-art baselines. \textbf{Bold} indicates best.} 
\begin{tabularx}{\linewidth}{lccc}
\midrule
  \textbf{Method} &  \textbf{ADE ($\text{m}$) $\downarrow$} & \textbf{FDE ($\text{m}$) $\downarrow$}& \textbf{AOE ($\text{rad}$) $\downarrow$}\\
\midrule
  ViNT & 0.756 $\pm$ 0.396 & 1.304 $\pm$ 0.788 & 0.203 $\pm$ 0.199 \\
  NoMaD & 0.853 $\pm$ 0.486 & 1.561 $\pm$ 0.979 & 0.251 $\pm$ 0.265 \\
  GNM & 0.753 $\pm$ 0.397 & 1.254 $\pm$ 0.778 & 0.219 $\pm$ 0.194 \\
  HST+MPC & 0.891 $\pm$ 0.825 & 1.886 $\pm$ 1.009 & 0.313 $\pm$ 0.210 \\
\rowcolor{gray!20} \textbf{\ours} & \textbf{0.578 $\pm$ 0.374} & \textbf{1.064 $\pm$ 0.703} & \textbf{0.159 $\pm$ 0.153} \\
\midrule
\end{tabularx} \vspace{-1.0em}
\label{table:quant}
\end{table}  \setlength{\tabcolsep}{6pt}

\subsection{Quantitative Results}

Table~\ref{table:quant} summarizes the comparative performance of each method.
\ours{} consistently outperforms all baselines across the three metrics. Compared to the strongest baseline on each metric, \ours{} reduces ADE by 23.2\%, FDE by 14.2\%, and AOE by 21.7\%. The gains in ADE and AOE suggest that our model better captures the subtle, socially aware maneuvers demonstrated by human teleoperators. The improved FDE is largely due to training with explicit goal positions. Vision-only baselines such as ViNT, NoMaD, and GNM condition on goal images, which can be visually ambiguous and lack geometric precision for long-horizon navigation, whereas \ours{} uses spatial coordinates that provide a more precise geometric objective.

Finally, while HST with MPC leverages high-fidelity human keypoints, its lower accuracy underscores a critical challenge. 
Integrating these forecasts into a decoupled MPC planner is inherently difficult, as the optimizer often struggles to balance goal-seeking with the rapidly shifting constraints of predicted keep-out zones. By learning social affordances in an end-to-end manner, \ours{} bypasses the sensitivity of these decoupled pipelines, resulting in more robust and socially compliant performance. 

\subsection{Qualitative Results}
Fig.~\ref{fig:qualitative} illustrates qualitative results by projecting predicted trajectories onto the robot's egocentric view. While quantitative analysis is conducted on out-of-distribution samples to ensure fairness, we also include results from SCAND test samples to demonstrate the results across diverse environmental contexts. 
We highlight several challenging social interaction scenarios to evaluate the model's behavior. 

We observe that purely vision-based baselines, such as ViNT, NoMaD, and GNM, suffer from significant performance degradation in scenarios with high scene variance, like involving turns. These baselines tend to simply maintain a straight trajectory regardless of the context, often failing to account for human presence. In contrast, our method, along with HST+MPC, remains strictly goal-oriented. HST+MPC often produces non-smooth trajectories because keep-out zones severely restrict feasible paths, causing abrupt steering and oscillations. 
This suggests that our approach is better suited for robot navigation, where providing a positional goal is more effective for consistent movement. Furthermore, our method better captures social compliance, proactively adjusting its path and timing to minimize interference with pedestrians while maintaining steady goal progress.



\setlength{\tabcolsep}{5.5pt}
\begin{table}[tb] 
\centering
\caption{\textbf{Ablation Study Results}: We compare the effect of individual components of \ours{}. \textbf{Bold} indicates best.} 
\begin{tabularx}{\linewidth}{lccc}
\midrule
  \textbf{Method} &  \textbf{ADE ($\text{m}$) $\downarrow$} & \textbf{FDE ($\text{m}$) $\downarrow$}& \textbf{AOE ($\text{rad}$) $\downarrow$}\\
\midrule
\textit{\ours-vision} & 0.585 $\pm$ 0.367 & 1.078 $\pm$ 0.692 & \textbf{0.157 $\pm$ 0.141} \\
\textit{\ours-hum} & 0.594 $\pm$ 0.377 & 1.095 $\pm$ 0.712 & {0.157 $\pm$ 0.143} \\
\rowcolor{gray!20} \textbf{\ours} & \textbf{0.578 $\pm$ 0.374} & \textbf{1.064 $\pm$ 0.703} & 0.159 $\pm$ 0.153 \\
\midrule
\end{tabularx} \vspace{-1.0em}
\label{table:abal}
\end{table}  \setlength{\tabcolsep}{6pt}

\subsection{Ablation Study Results}

To evaluate the contribution of different components in \ours{}, we perform two sets of ablation studies. 
First, we analyze the role of human whole-body pose cues to answer the question: \textit{``Is Body Language Actually Useful?"} 
Second, we analyze the effect of auxiliary human trajectory prediction to answer the question: \textit{``Does Explicitly Predicting Human Trajectories Help?"}

\vspace{5pt}\noindent {\bf Ablations on Keypoints:} We compare students distilled from our full Teacher model against those distilled from \textit{\ours-vision}, a teacher trained without skeletal input. As shown in Table~\ref{table:abal}, the model distilled from the skeleton-aware teacher achieves lower ADE and FDE while \textit{\ours-vision} yields a marginally better AOE. The overall differences are modest, which is expected given that the student operates purely from visual input at test time and has no direct access to skeletal cues. Nevertheless, the improvement in displacement metrics suggests that skeleton-aware teacher representations encode richer spatial social semantics that are partially transferable through distillation. The slightly higher AOE of \ours{} reflects more active heading adjustments when navigating around humans. \ours{} adjusts its heading more proactively, resulting in more human-aware but slightly less direct trajectories.

\vspace{5pt}\noindent {\bf Ablations on Human Trajectory Prediction:} We investigate whether supervising the Teacher model with an additional auxiliary human trajectory prediction objective improves navigation performance. In this configuration, a lightweight MLP head is added to predict future positions of each tracked human from the contextualized human tokens produced by the social-robot interaction stage. We name this variant \textit{\ours-hum}. As shown in Table~\ref{table:abal}, \textit{\ours-hum} causes performance to degrade by 2.8\% on average compared to our full model. We attribute this to the difficulty of explicitly predicting human trajectories, which is a distinct and non-trivial problem from robot trajectory planning. Jointly optimizing both objectives introduces conflicting gradient signals, ultimately hurting navigation performance. We therefore exclude it from our final model.

\subsection{Robot Deployment}
To validate the practicality of our approach, we deployed our model on a Clearpath Jackal equipped with a ZED2 RGB camera and NVIDIA GTX 1050 GPU. We compared our method against baseline methods in real-world scenarios. The navigation goal is fixed at a position $10\mathrm{m}$ ahead of the robot. We evaluate on two representative social navigation scenarios: frontal human approach and intersection crossing. 
In both scenarios, baseline methods tend to prioritize following a straight path toward the goal, often failing to yield to nearby humans in a timely manner. In contrast, \ours{} exhibits more conservative and human-aware behavior, proactively adjusting its path to avoid humans while maintaining progress toward the goal.

To further assess whether our model captures human motion intent, we test a scenario where a human approaches the robot diagonally, once from the right and once from the left. We observe that the robot consistently yields in the direction that minimizes interference with the approaching human: steering left when the human approaches from the right, and steering right when the human approaches from the left. This directionally-aware avoidance behavior suggests that \ours{} appears to infer human motion intent from visual cues alone at deployment time, without any access to explicit skeletal or tracking information. 
Despite these promising behaviors, our current policy is not yet robust enough for general-purpose open-world deployment.

%% file: content/5_conclusion.tex

\section{Conclusion}

In this paper, we present \ours, an end-to-end social robot navigation approach that leverages whole-body pose as an implicit cue for trajectory planning. We use a Teacher--Student strategy to distill privileged pose knowledge into a lightweight vision-only Student suitable for real-robot deployment, enabling socially compliant navigation without expensive skeletal tracking at runtime. Extensive comparisons against state-of-the-art baselines show that \ours\ achieves better social compliance, with an average gain of 29.8\%, demonstrating that implicit social awareness can be deployed on resource-constrained mobile platforms.


Despite these encouraging results, reliable real-world social navigation remains an open challenge. Although our policy exhibits promising behaviors in representative deployment scenarios, it is not yet robust enough for reliable deployment in unconstrained real-world environments. In particular, crowded scenes, occlusions, multiple interacting pedestrians, and unexpected human behaviors remain challenging. Furthermore, our approach is limited by the quality and diversity of available datasets. Since SCAND is collected through teleoperation, demonstrated trajectories may reflect operator preferences and interaction biases, making it difficult to learn broadly generalizable social behaviors.
Future work will focus on improving robustness through larger-scale, more diverse datasets with crowded, multi-person interactions under natural human behavior. 
We also plan to explore more effective representation learning and distillation strategies that capture complex social dynamics while remaining suitable for real-time deployment.

%% file: content/6_ack.tex
\section{Acknowledgement}



George Mason University is supported by the National Science Foundation (NSF 2350352), the Army Research Office (W911NF2320004, W911NF2520011), Google DeepMind, Clearpath Robotics, FrodoBots Lab, Raytheon Technologies, Tangenta, Mason Innovation Exchange, and Walmart.

Ewha Womans University is supported by the Institute of Information \& Communications Technology Planning \& Evaluation (IITP) grant funded by the Korean government (MSIT) (No. IITP-2026-RS-2020-II201460, Information Technology Research Center (ITRC)).


